\documentclass{article}

\usepackage{graphicx}

\usepackage{caption}

\usepackage{longtable}
\usepackage{multirow}

\usepackage{booktabs}
\usepackage[load-configurations=version-1]{siunitx} 

\usepackage{placeins}

\usepackage{url}

\title{Deep Similarity Learning Loss Functions in Data Transformation for Class Imbalance}

 \author{Damian Horna \\ Poznan University of Technology \\ {\tt horna.damian@gmail.com}     
 \and
  Mateusz Lango \\
   Poznan University of Technology, Institute of Computing Science,\\  60-965 Poznan, Poland\\
   Charles University, Faculty of Mathematics and Physics, \\ Prague, Czech Republic
   {\tt mlango@cs.put.poznan.pl} 
   \and
   Jerzy Stefanowski \\
   Poznan University of Technology, Institute of Computing Science,\\  60-965 Poznan, Poland 
   {\tt jstefanowski@cs.put.poznan.pl} }

\begin{document}

\maketitle

\begin{abstract}
Improving the classification of multi-class imbalanced data is more difficult than its two-class counterpart. In this paper, we use deep neural networks to train new representations of tabular multi-class data. Unlike the typically developed re-sampling pre-processing methods, our proposal modifies the distribution of features, i.e. the positions of examples in the learned embedded representation, and it does not modify the class sizes. To learn such embedded representations we introduced various definitions of triplet loss functions: the simplest one uses weights related to the degree of class imbalance, while the next proposals are intended for more complex distributions of examples and aim to generate a safe neighborhood of minority examples. Similarly to the resampling approaches, after applying such preprocessing, different classifiers can be trained on new representations. Experiments with popular multi-class imbalanced benchmark data sets and three classifiers showed the advantage of the proposed approach over popular pre-processing methods as well as basic versions of neural networks with classical loss function formulations.
\noindent This is an extended version of the paper presented at LIDTA'2023 workshop at ECMLPKDD2023 \cite{Horna2023}.

\end{abstract}

\section{Introduction}
\label{sec:intro}

Class imbalance is a common challenge in many domains, which severely degrades the predictive ability of classifiers. It has been extensively studied in the last decades and as a result, many methods have been proposed to improve classifiers, see e.g. \cite{Branco,Fernandez2018,Haixiang,krawczyk2016learning}. However, most of the proposed methods considered different modifications of the data or algorithms for fairly classic classifiers, such as k-NN, decision trees, SVM, ensembles or similar ones. Despite their effectiveness, especially for the processing of tabular data, the situation has begun to change with the development of deep neural networks. As noted by \cite{johnson2019survey}, one of the reasons for the success of deep learning is the ability to map the original data to an embedded new representation, where learning complex relationships can be much easier.

An increase in research interest in deep networks can also be observed in the class imbalance community, see e.g. \cite{buda2018,pmlr-v183-wojciechowski22a}. However, the number of such works is still insufficient. Moreover, most of them are concerned with the use of known preprocessing techniques in combination with selected neural network architectures, or with the modification of the error or loss functions to make them more sensitive to minority class errors.

In our opinion, the impact of learning a new embedded representation in the last layers of the network is still insufficiently exploited, especially for tabular data. It should be directed to obtaining better separation of minority classes examples from majority classes ones and linking this to learning further more efficient classifiers.  In this paper, we follow inspirations from the \textit{contrastive learning paradigm} \cite{jaishwal:2021}. The deep neural network is learned by comparing the feature representations of selected examples and directing the learning process of the network to find the differences and similarities between examples from different classes.

According to \cite{Schroff2015FaceNetAU} one can use \textit{triples} of examples and the corresponding contrast loss function, so-called \textit{triplet loss}, so that examples from the same class are brought closer to each other and at the same time are moved away from an example from another class in the embedded representation. Such an approach was mainly used to improve image recognition. Despite interesting extensions, such as  \cite{Huang2016LearningDR}, it has not been sufficiently explored so far for difficult imbalanced data. Therefore, in this paper,  we will study their usefulness for multi-class imbalanced data.

Recall that the multi-class imbalanced data are recognized as harder to learn than two-class ones due to more complex relationships between many classes and local difficulty characteristics for single examples from these classes \cite{lango2022makes}. Instead of looking for yet another method of solving such difficult multi-class problems in the original data representations, we will examine the possibility of learning new representations of these problems using appropriately modified triplet loss functions and, then, building classifiers in these embedded representations. 

The contributions of our paper are the following: 
\begin{enumerate}
\item to introduce modifications of the triplet loss functions to deal with imbalanced classes and to propose new ones to handle local difficulty of learning examples; 
\item  to present modifications of learning deep neural networks with them for processing multiple imbalanced tabular data; 
\item  to carry out the experimental analysis of proposed variants of these triplet loss functions and to compare the best of them against using previously known methods for dealing with multi-class imbalanced data.
\end{enumerate}

\section{Related works}
\label{sec:related}

The problems of learning from imbalanced data have been studied by many researchers - see e.g.  comprehensive reviews, such as \cite{Branco,Fernandez2018}. Most of the current research concerns binary imbalanced data, i.e. a single minority class versus the majority one \cite{japkowicz2016big}. Multi-class imbalanced classification is more difficult and less frequently studied than its binary counterpart. The specialized classification methods for this problem are usually divided into two groups: decomposition-based approaches (the transformation of the multi-class problem into a set of binary ones) and dedicated multi-class approaches (mainly various pre-processing methods). It is often postulated that such approaches should take into account both global class inter-relationships and local data difficulty factors \cite{saez2016analyzing,lango2022makes,Stefanowski2016}. 

Multi-class imbalanced classification is more difficult and less frequently studied than its binary counterpart. According to recent work, this may be due to many factors: overlapping between multiple classes, its interaction with various imbalance ratios, different configurations of class sizes and presence of the local difficulty distribution in the neighbourhoods of examples from particular classes \cite{lango2022makes,santos2022joint}.  The specialized classification methods for this problem are usually divided into two groups: decomposition-based approaches and dedicated multi-class approaches. The first methods rely on transformation of multi- class problems into a series of binary ones, which are solved separately and later aggregated in a kind of ensembles. The other category are usually data resampling techniques specialized for dealing with multiple classes, see e.g. random oversampling of classes, such as Global-CS \cite{GlobalCS:Zhou:2006} and more informed methods, e.g., Mahalanobis-distance over-sampling (MDO) \cite{MDO} or Static-SMOTE \cite{StaticSmote}. For their more comprehensive review see \cite{Fernandez2018,Grina}. 
Referring to the identified sources of difficulties in multiple imbalanced data  recall that the new methods designed for improving classifiers should take into account both both global class inter-relationships and local factors \cite{lango2017evaluating,lango2022makes,Krysia_2016}.

Despite advances in deep learning, multi-class imbalanced classification problems in the context of deep neural networks were mostly addressed for image classification or natural language processing tasks only; for their survey see \cite{buda2018,johnson2019survey}. Early proposals used resampling techniques before training deep neural networks. More recent works modified the distribution of examples in mini-batches during network training (see e.g.  ReMix algorithm by \cite{bellinger2018}) or oversampled the embedded representation. Another group of proposals is aimed at modifying the loss function in network training to make it sensitive to minority class errors, e.g. \cite{wang2016} modified the mean square error, \cite{focal} introduced focal loss, \cite{li-etal-2020-dice} proposed Dice loss, etc. Finally, there are also proposals of modifying the optimization task to handle different costs of misclassifications, see e.g. \cite{khan2017cost}. In addition, the more recent proposals try to preprocess the embedded representation,  such as DeepSmote \cite{dablain2021deepsmote} which uses the SMOTE algorithm on a hidden representation created with autoencoders or with GAN deep networks \cite{Salazar}. The results of these and other proposals are also discussed in \cite{johnson2019survey}.





In contrastive learning, the information about similarity or dissimilarity of training examples is used to learn corresponding feature representation. One of the most popular approaches is to train a neural network with \textit{triplet loss}, which has been introduced in \cite{Schroff2015FaceNetAU}. The method was originally applied to the face recognition task, where each face image was mapped to a compact Euclidean space, where distances directly corresponded to a measure of face similarity.  They used a deep convolutional neural network and trained it using triplets of matching / non-matching face patches generated using the triplet mining method. 

While using triplet loss, the neural network \textit{f}, which maps a training example $x \in X$ from the original feature space to the embedding space $f(x) \in R^d$, is trained to ensure that a selected reference example $x_i^a$ (called an \textit{anchor}) is closer to all other examples $x_i^p$ (\textit{positive}) from the same class than to any example $x_i^n$ (\textit{negative}) from other classes in the induced embedding space.  More formally, the constraint can be described as:

\begin{equation}
  \forall_{\left(f\left(x_{i}^{a}\right), f\left(x_{i}^{p}\right), f\left(x_{i}^{n}\right)\right) \in \mathcal{T}} \,\, \left\|f\left(x_{i}^{a}\right)-f\left(x_{i}^{p}\right)\right\|_{2}^{2}+\alpha<\left\|f\left(x_{i}^{a}\right)-f\left(x_{i}^{n}\right)\right\|_{2}^{2},
  \label{tripetConstraint}
\end{equation}
where $\alpha$ is a margin that should be enforced between positive and negative pairs and $\mathcal{T}$ is set of all possible triplets in the training set. In practice, this constraint leads to optimization of the following training objective:
\begin{equation}
    \sum_{i}^{|\mathcal{T}|}\left[\left\|f\left(x_{i}^{a}\right)-f\left(x_{i}^{p}\right)\right\|_{2}^{2}-\left\|f\left(x_{i}^{a}\right)-f\left(x_{i}^{n}\right)\right\|_{2}^{2}+\alpha\right]_{+}
    \label{tripletBasicObjective}
\end{equation}

The authors of \cite{Schroff2015FaceNetAU}  claim that his formulation of the loss function may lead to learning such embedded representations, where examples from the same class may approach each other in subsequent epochs of network learning, forming clusters. It worked very well in face recognitions and then it  proved to be successful especially while learning a similarity of images, in e.g., object tracking  or visual search. 
Such triplet loss functions were successfully applied in image recognition, object tracking, and visual search. Recently this idea has been modified to the quintuplet instance sampling and triple-header hinge loss, which allows learning more complex relationships between classes and their potential cluster in the embedded representations \cite{chen:2017,zhai:2020,Huang2016LearningDR}. 
To the best of our knowledge, the triplet loss and its extensions have not been used in the context of tabular, multi-class imbalanced data.


\section{Modifying triplet networks for learning representation of imbalanced data}
\label{sec:proposal}

The above-defined triplet loss, despite improving the separability of examples from different classes in the hidden embedded representation, does not directly take into account class imbalance or other local data difficulty factors (e.g., class overlapping, presence of small, rare groups of examples, decomposition of the minority class into rare sub-concepts, see \cite{Krysia_2016}). Therefore, in this paper we will present its extensions to address these issues.

The processing pipeline is as follows: the original tabular data is fed to a neural network (trained with the new proposed loss function) which transforms it to a potentially easier embedded representation, produced by the last layer of the network. Then, such preprocessed data (i.e. the constructed data representation) is used to train the final classifier. 
Note that the method modifies only the feature representation, combating the decline of classification performance on imbalanced data by addressing local data difficulty factors (i.e.~producing safer representation) and not by modifying the number of examples in particular classes.

We will use simple, feedforward neural networks -- being sufficient for processing tabular data, which are considered in this work. They consist of 3 layers with PReLU activations in between \cite{PReLU}. The numbers of units in each layer vary between datasets as we tune these individually per dataset. We use \textit{triplet loss} (defined in Eq. \ref{tripletBasicObjective}) as the baseline loss function. To optimize the objective, we use Adam optimizer. 

Selecting triplets for the optimization of the loss function is challenging since choosing all possible triplets is computationally infeasible. \cite{Schroff2015FaceNetAU} used an online sampling of triplets during the construction of \textit{mini-batches} and considering only such combinations of training examples that violate the constraint from Eq. \ref{tripetConstraint}.  As is still insufficient, the sampling of triplets is performed online with \textit{mini-batches}. 
Following inspirations from the experiments of \cite{Schroff2015FaceNetAU}, we will experimentally study three selection strategies: 
\begin{enumerate}
\item selecting the hardest negative instance for each positive pair -- select a negative example that is closest from the anchor in the mini-batch;
\item  random hard negative for each positive pair -- select one of the negatives that gives positive triplet loss value (i.e.~violates the constraint in Eq. (\ref{tripetConstraint}));
\item semi-hard negative for each positive pair -- allow to select one of the negatives that are further away from the anchor than the positive, but still localized within the margin $\alpha$.
\end{enumerate}

\subsection{Weighted triplet loss with global imbalances}

First, we propose a \textit{triplet weighting} modification of Eq. \ref{tripletBasicObjective} to better handle the minority classes. For each anchor example, $x_i^a$ we multiply the loss by inverse class cardinality normalized to 1 as shown in Eq. \ref{tripletWeighting}. As a result, the examples from minority classes provide bigger gradients for the neural network, than examples from the majority classes:

\begin{equation}
    \sum_{i}^{|\mathcal{T}|}\left[\left [ \left\|f\left(x_{i}^{a}\right)-f\left(x_{i}^{p}\right)\right\|_{2}^{2}-\left\|f\left(x_{i}^{a}\right)-f\left(x_{i}^{n}\right)\right\|_{2}^{2}+\alpha \right ] \times w_{x_i^a}\right]_{+},
    \label{tripletWeighting}
\end{equation}
where $\mathcal{T}$ is a set of all triplets, 
$ 
    w_{x_i^a} = \frac{|x \in X: y_x = y_{x_i^a}| ^{-1}}
    {\sum_{c_i}^{C} \left[\left|x \in X: y_x = c_i\right|^{-1}\right]},
$ 
$X$ denotes a set of all training examples, $y_x$ is a class of example $x$, and $C$ is a set of all the classes. 

\subsection{Dealing with the safe local neighborhood in loss functions}

This modification takes into account global imbalance ratios present in the data and, similarly to the original triplet loss, attempts to join examples from the given class into a single cluster in the embedded representation. Now we will also consider its variants oriented toward learning more complicated data structures. 

Firstly we want to clean the local neighborhood of examples.  If  the example  $x \in X$  is described in $d$-dimensional space $R^d$, $D$ is a distance measure (e.q. Euclidean) and $KNN(x)$ denotes \textit{k} nearest neighbors of $x$, then the new loss function is defined as:

\begin{equation}
  min\sum_x\left [ \sum _{x' \in KNN(x) : y_x=y_{x'}} D(x,x') - \sum _{x' \in KNN(x) : y_x \neq y_{x'}} min \left \{ D(x,x') - \alpha, 0 \right \} \right ],
  \label{basicFormulationOfANewMethod}
\end{equation}

\begin{equation}
    \alpha = \{
        \begin{array}{rr}
            \max \limits_{x' \in KNN(x):y_x=y_{x'}} D(x, x') + 1,& \text{if } | KNN(x):y_x=y_{x'}| > 0 \\
            1,              & \text{otherwise},
        \end{array}
    \label{newMethodAlpha}
\end{equation}

Intuitively, our objective (Eq. \ref{basicFormulationOfANewMethod}) is to minimize the sum of distances to examples from the same class in the nearest neighborhood of $x$ and maximize the sum of distances to examples from other classes in the nearest neighborhood of $x$. Note that the maximization of distances to examples from other classes does not go further than $\alpha$, which is dynamically set to be greater than the distance to the farthest instance of the same class in the neighborhood.  Moreover, using $KNN(x)$ requires changing the algorithm of constructing mini-batches: we select randomly $B$ examples from the training set and for each of those we find their $k$ nearest neighbors, which ultimately forms the mini-batch. The preliminary experiments indicated the value of $k$ to be equal to 20.  Batch size $B$ was equal to 16 or 32.

The next extensions are designed to exploit more information on the class distribution inside $k$-neighborhood, similarly to earlier studies on local safeness of the minority class examples \cite{LNS2017,Krysia_2016}. 
In the first extension of the above objective, the loss value is additionally multiplied by a weight  $w_x$ dependent on the ratio of examples from the same class in the neighborhood of the example $x$ (the less, the bigger is the weight). This is to make the model less biased towards majority classes and, in particular, focus its operation towards separating difficult (i.e.~unsafe) examples:

\begin{equation}
    w_x = (1 - \frac{|KNN(x) : y_x=y_{x'}|}{|KNN(x)|})
    \label{newMethodWeight}
\end{equation}

In order to make the above loss more robust to outliers and reduce the chance of overfitting, another proposed variant of the loss function uses the above weight but with a cut-off below a certain threshold.
Intuitively, after reaching a certain small ratio of examples from the same class (e.g. 0.3) in the neighborhood, we do not further increase the weight. 
The cutoff weight is denoted as $c_x$ instead of $w_x$ - see definition below.
\begin{equation}
    c_x = min\left \{ w_x, 0.7 \right \}
    \label{cutoffWeight}
\end{equation}

Finally, instead of minimizing the sum of distances to examples in the neighborhood  in the loss function Eq. \ref{basicFormulationOfANewMethod}, we try minimizing (or maximizing) the mean distance to examples from the same class (or accordingly other classes) in the nearest neighborhood:

\begin{equation}
    min\sum_x\left [ \frac{\sum _{x' \in KNN(x) : y_x=y_{x'}} D(x,x')}{|KNN(x) : y_x=y_{x'}|} - \frac{\sum _{x' \in KNN(x) : y_x \neq y_{x'}} min \left \{ D(x,x') - \alpha, 0 \right \}}{|KNN(x) : y_x \neq y_{x'}|} \right ] \times w_x
  \label{meanDistNewMethod}
\end{equation}

As starting from a random representation and optimizing it according to any loss function that uses KNN in the induced embedding space gives very poor results, we decided to learn the initial representation with the autoencoder and simple MSE loss. Then, we take the pre-trained encoder and further train it using one of the four variants of the loss function specified above.

\section{Experiments}
\subsection{An experimental setup}
\label{sec:experiments}

The aims of our experiments are the following: (1) to compare the newly proposed weighted loss function vs. the original triplet loss;  (2) to compare the performance of classifiers trained on a new, triplet-based representation versus the standard pre-processing methods for multiple imbalanced classes: Global-CS \cite{GlobalCS:Zhou:2006}, Static-SMOTE \cite{StaticSmote}, MDO \cite{MDO}; (3) to visualize the embedding space of imbalanced data learned by triplet networks; (4) to evaluate usefulness of different variants of the proposed triplet loss functions.


In all these experiments we use KNN, Decision Tree, and LDA as final classifiers -- implementations were taken from the scikit-learn library. LDA (Linear Discriminant Analysis) is run with default parameters  and  KNN classifier with \textit{k} equal to 1. Decision Trees are parameterized with min\_samples\_split=4, min\_samples\_leaf=2, and class\_weight='balanced'. For Global-CS, Static-SMOTE or MDO, we used implementations provided by the multi -- imbalance library \cite{Grycza2020multiimbalanceOS}. Before applying any preprocessing, classification, or representation learning we performed one-hot encoding of the categorical attributes (typical for processing in scikit learn) as well as z-score standardization for numerical attributes. 

All experiments are carried out with 17 popular multi-class imbalanced benchmark datasets characterized by different levels of difficulty \cite{lango2022makes}. Due to space limits, we skip the table with their detailed characteristics, the reader can find their details e.g. in \cite{janicka}.

We decided to use \textit{F1-Score} (macro-averaged) and \textit{G-Mean} measures in their versions for multiple classes due to their popularity in the literature (for their definitions see \cite{Brzezinski2018,blaszczynski2013extending,Fernandez2018}) and estimate their values with the stratified 5-fold cross-validation technique.

\begin{table}[hbt]
\footnotesize
  \begin{center}
 \begin{tabular}{lrr}
  \toprule
\textbf{dataset}   & \textbf{unweighted} & \textbf{weighted} \\
  \midrule
  cmc        &  0.504  & \textbf{0.518}   \\
  dermatology   & 0.958    & \textbf{0.970}  \\
 hayes-roth      &  0.858    &  \textbf{0.864}  \\
vehicle            & 0.979      & \textbf{0.985}   \\
 yeast              &  \textbf{0.383}  &  0.294  \\
art1                 & 0.973                & \textbf{0.986}     \\
art2       &  0.779           &  \textbf{0.806}   \\
art3       & 0.557             & \textbf{0.589}      \\
art4       &  0.860           &  \textbf{0.880}  \\
balance-scale  & 0.887 & \textbf{0.969}     \\
 cleveland  &  0.136      &  \textbf{0.152}   \\
cleveland\_v2    & 0.192  & \textbf{0.211}  \\
 glass      &  0.646           &  \textbf{0.693}   \\
ecoli        & 0.814            & \textbf{0.817}    \\
 led7digit  &  \textbf{0.758}  &  0.757    \\
winequality-red   & 0.445     & \textbf{0.472}   \\
 thyroid    &  0.944  &  \textbf{0.951} \\
   \bottomrule
  \end{tabular}
 \caption{LDA \textit{G-Mean} results comparison for representations obtained with weighted and unweighted triplets}
\label{lda-gmean-triplet-weighting-comparison}
  \end{center}
\end{table}

\subsection{Studying weighted vs. unweighted triplets}

In the first experiment, we compare the usefulness of the triplet weighting, deined in Eq. \ref{tripletWeighting} with the original, unweighted objective ( Eq. \ref{tripletBasicObjective}). For brevity, we present the exact results only for the LDA classifier and G-mean measure in Table \ref{lda-gmean-triplet-weighting-comparison} - the best results are in bold.

An improvement of classification performance with the representation obtained with triplet weighting (described in Eq. \ref{tripletWeighting}) is clearly visible. The representation obtained through the original objective \ref{tripletBasicObjective} provides better results on the \textit{G-Mean} only for two datasets: yeast and led7digit. Similar observations are done for the F1 measure. We also compared all sampling strategies for choosing triplets into the mini-batch in training a neural network. Results showed that the differences between them are rather small. Due to small computational costs, we stick to the random hard negative strategy in the rest of experiments.


\subsection{Comparison of pre-processing methods vs. new learned representation}

Then we compare the performance of all three classifiers learned on the new embedded representation (obtained with the weighted triplets - Eq. \ref{tripletBasicObjective}) against using these classifiers with specialized preprocessing methods Global-CS, Static-SMOTE and MDO (their description see \cite{janicka}). Additionally, we will learn these classifiers directly on the original, unchanged data - which is denoted as Baseline.  Again for brevity, we present the results only for one selected classifier: KNN, see Tables \ref{knn-f1-comparison} and \ref{knn-Gmean-comparison}, respectively.

\begin{table}[hbt]
\footnotesize
  \begin{center}
 \begin{tabular}{lrrrrr}
  \toprule
\textbf{dataset} & \textbf{Baseline} & \textbf{Global-CS} & \textbf{Static-SMOTE}  & \textbf{MDO}   & \textbf{New Rep.}      \\
  \midrule
cmc              & 0.423 ± 0.015     & 0.409 ± 0.008      & 0.415 ± 0.004          & 0.415 ± 0.008          & \textbf{0.454 ± 0.008} \\
dermatology      & 0.937 ± 0.018     & 0.891 ± 0.021      & 0.910 ± 0.040          & 0.891 ± 0.020          & \textbf{0.970 ± 0.029} \\
hayes-roth       & 0.672 ± 0.113     & 0.817 ± 0.053      & 0.823 ± 0.059          & 0.834 ± 0.060          & \textbf{0.886 ± 0.042} \\
vehicle          & 0.912 ± 0.023     & 0.912 ± 0.023      & 0.909 ± 0.036          & 0.912 ± 0.023          & \textbf{0.989 ± 0.021} \\
yeast            & 0.487 ± 0.039     & 0.487 ± 0.039      & 0.458 ± 0.026          & 0.467 ± 0.036          & \textbf{0.497 ± 0.045} \\
art1             & 0.946 ± 0.012     & 0.946 ± 0.012      & 0.950 ± 0.013          & 0.946 ± 0.017          & \textbf{0.974 ± 0.015} \\
art2             & 0.700 ± 0.035     & 0.700 ± 0.035      & 0.702 ± 0.023          & 0.712 ± 0.012          & \textbf{0.751 ± 0.032} \\
art3             & 0.509 ± 0.041     & 0.509 ± 0.041      & 0.479 ± 0.033          & 0.528 ± 0.049          & \textbf{0.572 ± 0.073} \\
art4             & 0.768 ± 0.014     & 0.768 ± 0.014      & 0.769 ± 0.020          & 0.781 ± 0.034          & \textbf{0.817 ± 0.017} \\
balance-scale    & 0.561 ± 0.009     & 0.345 ± 0.044      & 0.304 ± 0.047          & 0.345 ± 0.041          & \textbf{0.954 ± 0.040} \\
cleveland        & 0.297 ± 0.042     & 0.317 ± 0.073      & 0.300 ± 0.043          & \textbf{0.328 ± 0.078} & 0.315 ± 0.105          \\
cleveland\_v2    & 0.327 ± 0.064     & 0.322 ± 0.037      & \textbf{0.332 ± 0.058} & 0.328 ± 0.025          & 0.306 ± 0.062          \\
glass            & 0.659 ± 0.065     & 0.659 ± 0.065      & 0.673 ± 0.071          & 0.666 ± 0.064          & \textbf{0.690 ± 0.080} \\
ecoli            & 0.793 ± 0.054     & 0.793 ± 0.054      & 0.776 ± 0.075          & 0.800 ± 0.062          & \textbf{0.814 ± 0.049} \\
led7digit        & 0.734 ± 0.035     & 0.706 ± 0.040      & 0.681 ± 0.044          & 0.706 ± 0.035          & \textbf{0.748 ± 0.048} \\
winequality-red  & 0.535 ± 0.029     & 0.535 ± 0.029      & 0.516 ± 0.038          & 0.537 ± 0.035          & \textbf{0.553 ± 0.055} \\
thyroid          & 0.930 ± 0.042     & 0.930 ± 0.042      & 0.937 ± 0.042          & 0.930 ± 0.042          & \textbf{0.957 ± 0.034}  \\
 \bottomrule
  \end{tabular}
 \caption{Comparison of \textit{F1-Score} for KNN learned on the original data (Baseline), after using pre-processing methods and for a new weighted triplet representation}
\label{knn-f1-comparison}
\end{center}
\end{table}

\begin{table}[hbt]
\footnotesize
  \begin{center}
 \begin{tabular}{lrrrrr}
  \toprule
\textbf{dataset} & \textbf{Baseline} & \textbf{Global-CS} & \textbf{Static-SMOTE}  & \textbf{MDO}   & \textbf{New Rep.}      \\
  \midrule
cmc              & 0.413 ± 0.017     & 0.404 ± 0.008      & 0.408 ± 0.007          & 0.412 ± 0.009          & \textbf{0.448 ± 0.012} \\
dermatology      & 0.932 ± 0.024     & 0.892 ± 0.022      & 0.916 ± 0.035          & 0.892 ± 0.022          & \textbf{0.968 ± 0.031} \\
hayes-roth       & 0.636 ± 0.114     & 0.803 ± 0.059      & 0.812 ± 0.066          & 0.824 ± 0.065          & \textbf{0.881 ± 0.045} \\
vehicle          & 0.912 ± 0.018     & 0.912 ± 0.018      & 0.914 ± 0.032          & 0.912 ± 0.018          & \textbf{0.990 ± 0.021} \\
yeast            & 0.408 ± 0.088     & 0.408 ± 0.088      & \textbf{0.415 ± 0.080} & 0.364 ± 0.099          & 0.405 ± 0.099          \\
art1             & 0.944 ± 0.012     & 0.944 ± 0.012      & 0.952 ± 0.019          & \textbf{0.966 ± 0.017} & 0.964 ± 0.029          \\
art2             & 0.678 ± 0.052     & 0.678 ± 0.052      & 0.706 ± 0.027          & 0.728 ± 0.015          & \textbf{0.735 ± 0.033} \\
art3             & 0.458 ± 0.049     & 0.458 ± 0.049      & 0.470 ± 0.046          & 0.526 ± 0.061          & \textbf{0.541 ± 0.104} \\
art4             & 0.761 ± 0.020     & 0.761 ± 0.020      & 0.784 ± 0.024          & 0.802 ± 0.046          & \textbf{0.811 ± 0.029} \\
balance-scale    & 0.089 ± 0.001     & 0.064 ± 0.005      & 0.056 ± 0.006          & 0.065 ± 0.005          & \textbf{0.936 ± 0.065} \\
cleveland        & 0.078 ± 0.041     & 0.165 ± 0.158      & 0.121 ± 0.043          & 0.174 ± 0.166          & \textbf{0.180 ± 0.156} \\
cleveland\_v2    & 0.065 ± 0.042     & 0.048 ± 0.034      & \textbf{0.116 ± 0.044} & 0.049 ± 0.033          & 0.056 ± 0.035          \\
glass            & 0.554 ± 0.183     & 0.554 ± 0.183      & \textbf{0.589 ± 0.196} & 0.563 ± 0.186          & 0.573 ± 0.207          \\
ecoli            & 0.773 ± 0.058     & 0.773 ± 0.058      & 0.769 ± 0.075          & \textbf{0.799 ± 0.067} & 0.786 ± 0.067          \\
led7digit        & 0.648 ± 0.207     & 0.655 ± 0.051      & 0.647 ± 0.060          & 0.571 ± 0.182          & \textbf{0.699 ± 0.089} \\
winequality-red  & 0.475 ± 0.043     & 0.475 ± 0.043      & 0.488 ± 0.045          & \textbf{0.495 ± 0.048} & 0.453 ± 0.094          \\
thyroid          & 0.917 ± 0.080     & 0.917 ± 0.080      & 0.926 ± 0.080          & 0.917 ± 0.080          & \textbf{0.937 ± 0.061} \\
 \bottomrule
  \end{tabular}
 \caption{Comparison of \textit{G-mean} for KNN learned on the original data (Baseline), after using pre-processing methods and for a new weighted triplet representation}
\label{knn-Gmean-comparison}
\end{center}
\end{table}

\begin{table}[h]
\footnotesize
\centering
\begin{tabular}{lrrrrr}
\hline
             & \textbf{New Rep.} & \textbf{MDO} & \textbf{Baseline} & \textbf{Static-SMOTE} & \textbf{Global-CS} \\\hline
\textbf{KNN} & 1.35              & 2.71         & 3.47              & 3.65                  & 3.82               \\
\textbf{DT}  & 1.59              & 2.94         & 2.94              & 3.65                  & 3.88               \\
\textbf{LDA} & 1.41              & 3.59         & 3.82              & 3.65                  & 2.53              \\\hline
\end{tabular}
\caption{Average ranks of compared algorithms (the lower, the
better) from Friedman tests on \textit{F1-Score} for all  classifiers.}
\label{f1-friedman}
\end{table}

\begin{table}[h]
\footnotesize
\centering
\begin{tabular}{lrrrrr}\hline
             & \multicolumn{1}{l}{\textbf{New Rep.}} & \multicolumn{1}{l}{\textbf{MDO}} & \multicolumn{1}{l}{\textbf{Baseline}} & \multicolumn{1}{l}{\textbf{Static-SMOTE}} & \multicolumn{1}{l}{\textbf{Global-CS}} \\\hline
\textbf{KNN} & 1.71                                  & 2.85                             & 3.59                                  & 2.88                                      & 3.97                                   \\
\textbf{DT}  & 1.82                                  & 3.00                             & 3.00                                  & 2.94                                      & 4.24                                   \\
\textbf{LDA} & 1.59                                  & 3.65                             & 4.24                                  & 3.53                                      & 2.00            \\\hline                      
\end{tabular}
\caption{Average ranks of compared algorithms (the lower, the
better) from Friedman tests on \textit{G-Mean} for
various classifiers.}
\label{gmean-friedman}
\end{table}

Results for both measures (also for Decision Trees and LDA) show that using the new embedded representation leads to better classifier performance than the compared pre-processing methods. We also carried out the Friedman ranked test and rejected the null hypothesis of similar performance of compared methods. The averaged ranks for all the classifiers are given in Tables \ref{f1-friedman} and \ref{gmean-friedman} for F1 and G-mean , respectively. One can notice that our newly proposed triplet representations are always better than the preprocessing methods, while MDO or GlobalCS are in the second position. The post-hoc Nemenyi analysis shows clear dominance for K-NN while for other classifiers the rank differences are within CD. However, according to the paired Wilcoxon test the new representation is significantly better for all classifiers in the case of F1 score. Additionally, the performance of the proposed approach is also statistically significantly better for KNN and DT with respect to G-mean (for LDA $p$ = 0.9).

\subsection{Visualization of the learned embeddings }
\label{subsec:visual}

To examine the changes in the example's distribution inside the embedded representation, we visualized them with PCA methods for two dimensions and compared them to the analogous visualization of the distribution in the original data. Below two representative datasets (easier and more difficult ones) are presented in Figures \ref{vehicle-original} -- \ref{ecoli-trained}. 

One can notice that easier datasets such as vehicle -- but also dermatology, hayes-roth or thyroid, which we were unable to include due to page limits - are easily transformed into single clusters of examples for each class. As they become separable in the embedded representation it somehow solved the overlapping in those datasets. Ecoli was a more complex data (see also its analysis in \cite{LNS2017}) and its separation in the embedded representation is not so clear as before, but there is still an improvement in the differentiation of examples from different classes. However, the situation is more difficult for few harder datasets, such as glass or led7digit. For these datasets, the method has difficulties in transforming the space into single clusters for each class. Those datasets form much more complicated structures in the original feature space (see experimental studies such as \cite{lango2022makes}), which makes them very hard to transform by the triplet-based network using only weights referring to the global imbalance. On the other hand the results on the balance-scale --  see Figures \ref{balance-scale-original} and \ref{balance-scale-trained} -- came as a surprise, because this dataset seemed hard to classify with KNN, Decision Tree or LDA on original features, but it was easily decomposed into single manifolds by the triplet-based network. This may suggest, that non-linear feature transformations found by the network can reduce difficulty factors even for some harder datasets.

\begin{minipage}{\linewidth}
    \begin{minipage}{\linewidth}
            \centering
            \includegraphics[width=5.8cm]{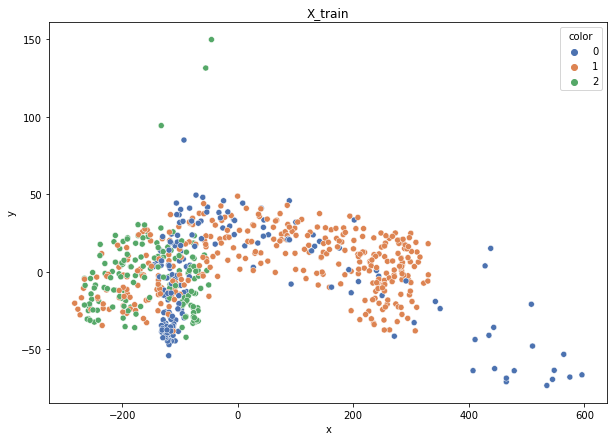}
            \captionof{figure}{Visualization of the original vehicle dataset in 2 PCA dimensions}
            \label{vehicle-original}
    \end{minipage}
    \begin{minipage}{\linewidth}
            \centering
            \includegraphics[width=5.8cm]{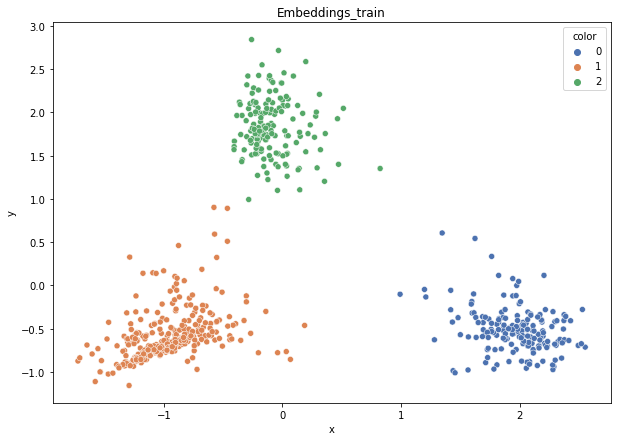}
            \captionof{figure}{Visualization of the triplet learned vehicle embeddings in 2 PCA dimensions}
            \label{vehicle-trained}
    \end{minipage}
\end{minipage}

\begin{minipage}{\linewidth}
    \begin{minipage}{\linewidth}
            \centering
            \includegraphics[width=5.8cm]{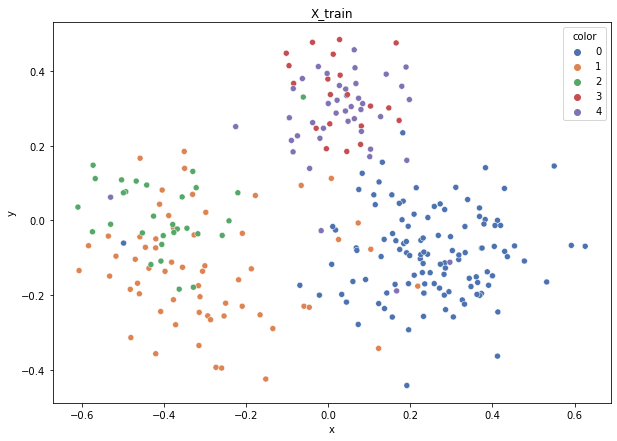}
            \captionof{figure}{Visualization of the original ecoli dataset in 2 PCA dimensions}
            \label{ecoli-original}
    \end{minipage}
    \begin{minipage}{\linewidth}
            \centering
            \includegraphics[width=5.8cm]{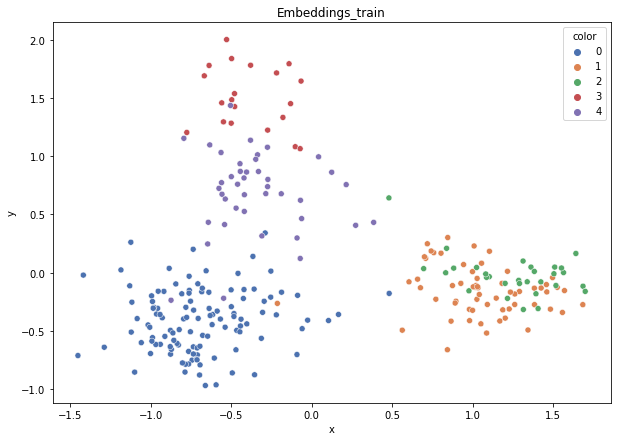}
            \captionof{figure}{Visualization of the triplet learned ecoli embeddings in 2 PCA dimensions}
            \label{ecoli-trained}
    \end{minipage}
\end{minipage}

\begin{minipage}{\linewidth}
    \begin{minipage}{\linewidth}
            \centering
            \includegraphics[width=6cm]{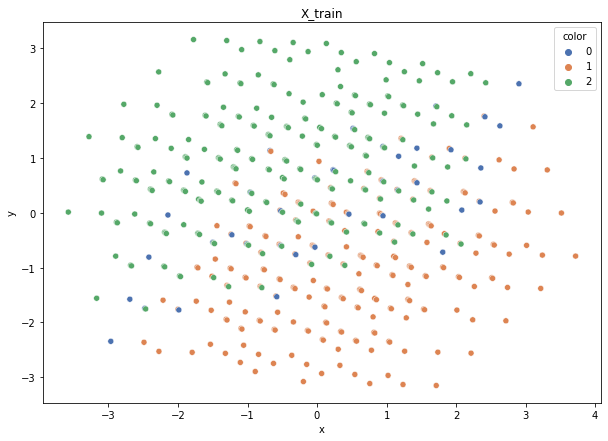}
            \captionof{figure}{Visualization of the original balance-scale dataset in 2 PCA dimensions}
            \label{balance-scale-original}
    \end{minipage}
    \begin{minipage}{\linewidth}
            \centering
            \includegraphics[width=6cm]{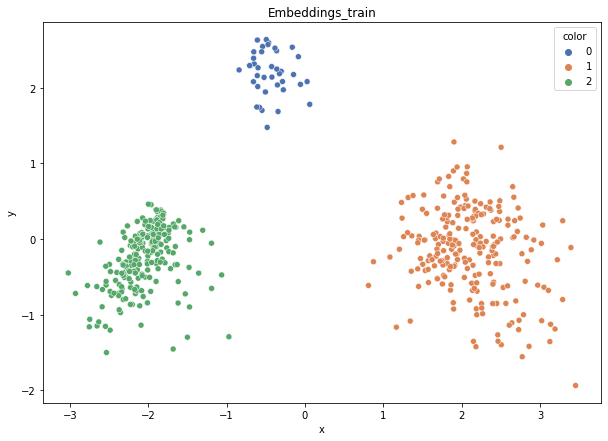}
            \captionof{figure}{Visualization of the triplet balance-scale embeddings in 2 PCA dimensions}
            \label{balance-scale-trained}
    \end{minipage}
\end{minipage}

\subsection{Experiments with new versions of triplet loss functions}

\begin{table}[hbt]
\tiny
  \begin{center}
 \begin{tabular}{lrrrrrr}
  \toprule
  \textbf{dataset} & \textbf{Baseline} & \textbf{Autoencoder} & \textbf{Basic}         & \textbf{Safe Weights} & \textbf{Cutoff}        & \textbf{Mean Dists}  \\
  \midrule
cmc   & 0.423 ± 0.015     & 0.403 ± 0.022        & 0.426 ± 0.014          & 0.430 ± 0.023             & 0.421 ± 0.011          & \textbf{0.436 ± 0.022} \\
dermatology & 0.937 ± 0.018     & 0.920 ± 0.017        & 0.946 ± 0.030          & 0.949 ± 0.017             & 0.946 ± 0.020   & \textbf{0.953 ± 0.023} \\
hayes-roth   & 0.672 ± 0.113     & 0.807 ± 0.051        & 0.880 ± 0.054          & \textbf{0.896 ± 0.051}   & 0.891 ± 0.045      & 0.891 ± 0.045   \\
vehicle         & 0.912 ± 0.023     & 0.918 ± 0.035        & 0.937 ± 0.010          & 0.975 ± 0.017            & 0.967 ± 0.022       & \textbf{0.986 ± 0.012} \\
yeast            & 0.487 ± 0.039     & 0.488 ± 0.040       & 0.488 ± 0.063          & 0.486 ± 0.050      & 0.487 ± 0.054          & \textbf{0.497 ± 0.052} \\
art1             & 0.946 ± 0.012     & 0.945 ± 0.012        & 0.945 ± 0.017          & 0.949 ± 0.024         & 0.947 ± 0.016          & \textbf{0.955 ± 0.021} \\
art2             & 0.700 ± 0.035     & 0.694 ± 0.024        & 0.714 ± 0.031          & 0.717 ± 0.026         & 0.711 ± 0.016          & \textbf{0.719 ± 0.038} \\
art3             & 0.509 ± 0.041     & 0.521 ± 0.037        & 0.523 ± 0.033          & 0.523 ± 0.026             & 0.494 ± 0.061      & \textbf{0.524 ± 0.036} \\
art4             & 0.768 ± 0.014     & 0.767 ± 0.016        & 0.778 ± 0.019          & 0.777 ± 0.028             & 0.765 ± 0.018       & \textbf{0.786 ± 0.018} \\
balance-scale    & 0.561 ± 0.009     & 0.401 ± 0.059        & 0.508 ± 0.028          & 0.839 ± 0.058        & 0.862 ± 0.054    & \textbf{0.942 ± 0.043} \\
cleveland        & 0.297 ± 0.042     & 0.321 ± 0.067        & \textbf{0.356 ± 0.083} & 0.319 ± 0.039         & 0.325 ± 0.034    & 0.335 ± 0.044       \\
cleveland\_v2    & 0.327 ± 0.064     & 0.310 ± 0.045        & 0.315 ± 0.025          & 0.314 ± 0.037        & 0.309 ± 0.015     & \textbf{0.342 ± 0.024} \\
glass            & 0.659 ± 0.065     & 0.642 ± 0.105        & 0.679 ± 0.035          & 0.690 ± 0.058        & 0.694 ± 0.056          & \textbf{0.699 ± 0.073} \\
ecoli            & 0.793 ± 0.054     & 0.776 ± 0.053        & 0.809 ± 0.048          & 0.789 ± 0.037         & \textbf{0.810 ± 0.036} & 0.802 ± 0.045          \\
led7digit        & 0.734 ± 0.035     & 0.708 ± 0.041        & \textbf{0.755 ± 0.076} & 0.748 ± 0.024       & 0.737 ± 0.079          & 0.744 ± 0.041          \\
winequality-red  & 0.535 ± 0.029     & 0.530 ± 0.027        & 0.536 ± 0.055          & \textbf{0.543 ± 0.030}    & 0.542 ± 0.033    & 0.537 ± 0.027     \\
thyroid          & 0.930 ± 0.042     & 0.946 ± 0.052    & 0.952 ± 0.053     & \textbf{0.953 ± 0.038}    & \textbf{0.953 ± 0.038} & \textbf{0.953 ± 0.038}  \\
 \bottomrule
  \end{tabular}
 \caption{Comparison of \textit{F1-Score} for KNN learned on the original data (Baseline), and for new versions of loss functions}
\label{knn-f1-comparison-new-method}
\end{center}
\end{table}

Finally, we tested all proposed new formulations of loss functions proposed in Section \ref{sec:proposal} -- see Eq. \ref{basicFormulationOfANewMethod} -- \ref{meanDistNewMethod}. Let us remind, that they are motivated by the observation that data can form very complicated structures in the feature space, which are not necessarily possible to represent as single clusters of examples for each class by using the simpler modification of the triplet loss function to the imbalance weighting - see also visualization results.

 As mentioned before, we first train an autoencoder to minimize the MSE loss. This is to avoid the collapse of training since the proposed loss functions depend on kNN in the embedding space which is random at the beginning of training. Therefore, after the autoencoder converges, we take the encoder network and continue training under one of the loss functions specified in Section \ref{sec:proposal}. They are denoted  as:
\begin{itemize}
    \item Basic -- triplet loss guided to clean example's neighborhood -- Eq. \ref{basicFormulationOfANewMethod}.
    \item Safeness Weights -- basic with examples' weights depending on their safe levels as defined in Eq. \ref{newMethodWeight}.
    \item Cutoff - another variant with safeness-based examples' weights but with additional cut-off at the threshold value of 0.7 as defined in Eq. \ref{cutoffWeight}.
    \item Mean Dists - the last variant analysing the mean distances to the examples from the same class as it is defined in Eq. \ref{meanDistNewMethod}.
\end{itemize}
In our experiments, we also calculated the results of classification performed on the original representation (this is our \textit{Baseline}) and on the representation obtained from the hidden layer of autoencoder to see, how much information is lost by such training initialization. 

\begin{table}[hbt]
\tiny
\centering
\begin{tabular}{lrrrrrr}
  \toprule
\textbf{dataset} & \textbf{Baseline}      & \textbf{Autoencoder} & \textbf{Basic}         & \textbf{Safeness Weights} & \textbf{Cutoff}        & \textbf{Mean Dists}    \\
  \midrule 
cmc    & 0.413 ± 0.017          & 0.398 ± 0.027        & 0.419 ± 0.015          & 0.424 ± 0.025             & 0.417 ± 0.014          & \textbf{0.429 ± 0.025} \\ 
dermatology      & 0.932 ± 0.024    & 0.927 ± 0.016  & 0.944 ± 0.027     & 0.950 ± 0.022             & 0.947 ± 0.023          & \textbf{0.953 ± 0.028} \\
hayes-roth       & 0.636 ± 0.114      & 0.793 ± 0.058        & 0.875 ± 0.058          & \textbf{0.892 ± 0.054}    & 0.886 ± 0.048          & 0.886 ± 0.048 \\
vehicle     & 0.912 ± 0.018          & 0.918 ± 0.031        & 0.935 ± 0.012          & 0.976 ± 0.017             & 0.967 ± 0.023          & \textbf{0.986 ± 0.013} \\
yeast  & \textbf{0.408 ± 0.088} & 0.407 ± 0.086        & 0.376 ± 0.121          & 0.375 ± 0.110             & 0.375 ± 0.111          & 0.385 ± 0.114          \\
art1    & 0.944 ± 0.012          & 0.942 ± 0.011        & 0.946 ± 0.017          & 0.952 ± 0.029             & 0.949 ± 0.019          & \textbf{0.956 ± 0.017} \\
art2    & 0.678 ± 0.052          & 0.676 ± 0.033        & 0.693 ± 0.032          & 0.710 ± 0.033             & 0.695 ± 0.016          & \textbf{0.712 ± 0.047} \\
art3    & 0.458 ± 0.049          & 0.474 ± 0.048        & 0.478 ± 0.031          & 0.478 ± 0.035             & 0.431 ± 0.062          & \textbf{0.479 ± 0.034} \\ 
art4    & 0.761 ± 0.020          & 0.764 ± 0.022        & 0.767 ± 0.030          & 0.774 ± 0.032             & 0.766 ± 0.029          & \textbf{0.781 ± 0.016} \\
balance-scale    & 0.089 ± 0.001  & 0.071 ± 0.008  & 0.158 ± 0.166        & 0.827 ± 0.080             & 0.855 ± 0.076          & \textbf{0.958 ± 0.037} \\
cleveland   & 0.078 ± 0.041  & 0.129 ± 0.134        & 0.161 ± 0.145          & 0.141 ± 0.089             & 0.146 ± 0.093          & \textbf{0.194 ± 0.127} \\
cleveland\_v2    & 0.065 ± 0.042          & 0.049 ± 0.038        & 0.056 ± 0.030          & 0.069 ± 0.030             & 0.053 ± 0.028          & \textbf{0.071 ± 0.025} \\
glass     & 0.554 ± 0.183     & 0.490 ± 0.262        & 0.561 ± 0.185          & 0.577 ± 0.196             & \textbf{0.642 ± 0.064} & 0.591 ± 0.214          \\
ecoli         & 0.773 ± 0.058   & 0.739 ± 0.057        & \textbf{0.797 ± 0.060} & 0.772 ± 0.034             & 0.793 ± 0.037          & 0.774 ± 0.051          \\
led7digit        & 0.648 ± 0.207          & 0.667 ± 0.078    & \textbf{0.743 ± 0.218} & 0.736 ± 0.028          & 0.647 ± 0.231          & 0.735 ± 0.052     \\
winequality-red  & 0.475 ± 0.043          & 0.460 ± 0.028        & 0.468 ± 0.089          & \textbf{0.493 ± 0.036}    & 0.492 ± 0.038          & 0.481 ± 0.041   \\
thyroid          & 0.917 ± 0.080          & 0.936 ± 0.069        & 0.937 ± 0.089          & \textbf{0.937 ± 0.058}    & \textbf{0.937 ± 0.058} & \textbf{0.937 ± 0.058} \\
 \bottomrule
\end{tabular}
\caption{KNN \textit{G-Mean} results comparison for different variants of the new loss function}
\label{knn-gmean-comparison-new-method}
\end{table}

The results for k-NN classifiers (with $k=1$) are presented in Table \ref{knn-f1-comparison-new-method}.  They show that the proposed method with new loss functions outperforms \textit{Baseline} classification on original representation and also for the autoencoder network. The comparison of different loss function variants clearly showed the advantage of the most advanced version, i.e. Mean Dists (which wins in 12 datasets). Versions that use weighting (i.e. Safeness Weights and Cutoff) are second in order of wins. A fairly similar dominance of the Mean Dists function is also evident for the G-mean measure (see Table \ref{knn-gmean-comparison-new-method}), although some advantages occur for other datasets.

However, comparing this best version of the new loss function to the first proposed function (weighted triplet as defined by Eq. \ref{tripletWeighting}) indicates a less significant advantage. According to the paired Wilcoxon signed test, the $p$ values are equal to 0.029 for F1-Score and 0.589 for G-mean (i.e. here the difference is not statistically significant). Looking more closely at the results, the better performance of more advanced versions of the loss function occurs for some datasets such as hayes-roth, balance scale, cleveland, glass, ecoli, led7digit lub wine-quality. Following previous analyses of their difficulties and complex distribution many of these datasets are identified as more difficult to learn than dermatology, vehicles or artificial datasets art \cite{lango2022makes}.

\section{Final remarks}

We presented a new way to create an embedded representation for multi-class imbalanced data that uses neural network training with specialized triplet loss functions. It should be emphasized that our proposal only modifies the distribution of features, i.e. the positions of examples in the newly learned deep representation, and does not modify the class size or does not create any synthetic examples (which is the most common for the currently most popular under/oversampling processing methods).

In accordance with the inspirations of contrastive learning in deep networks, we want to learn such a new representation of examples that their distribution of classes in the new embedded space becomes more separated and safe in the sense of the neighborhood of examples from a given class \cite{Krysia_2016} and we do not change the degree of imbalance in the dataset.    

For this purpose, we introduced various definitions of triplet loss functions used in training selected deep neural networks for tabular data. The first and simplest proposal modifies the original triplet loss function \cite{Schroff2015FaceNetAU} by introducing weights related to the degree of class imbalance. As can be seen from the conducted experiments (see, for example, visualizations in section \ref{subsec:visual}), this can lead to learning disjoint clusters of objects from different classes. Next, we put forward completely new proposals of triplet loss functions which are intended for more complex distributions of examples, for which it is necessary to take into account the local degrees of safeness of the examples against their nearest neighbors.



Quite comprehensive experiments with popular benchmark multi-class imbalanced data sets and three classifiers -- K-NN, decision trees and linear discriminant analysis --  clearly showed the advantage of the proposed approach over the use of popular pre-processing methods as well as basic versions of neural networks with classic loss function formulations. However, it should be also noted that for many of the analyzed datasets, the differences between the proposed loss function variants were not statistically significant. Nevertheless, the most advanced version, i.e. triplet loss with mean distances (Mean Dist)  was by far the winner of most comparisons, but for some datasets the weighted triplet loss function (as defined by Eq. \ref{tripletWeighting}) worked better. For additional results please refer to our external repository:\\  
\url{https://github.com/damian-horna/imbalanced_triplets}.

Despite encouraging experimental results, we believe that in future research it is still worth considering using and possibly extending the recent modifications of the triplet functions. For instance, one can start by investigating the loss functions considering the positions of four  (inspired by the work on other loss functions with four elements \cite{chen:2017}) or even five raining instances instead of triples \cite{Huang2016LearningDR}, since they could potentially allow for the creation of even more complicated representations for particularly difficult imbalanced data distributions. Another line of research is to test the applicability of the new triplet loss functions also for binary classification. Preliminary ongoing experiments have already shown their effectiveness.

\noindent \textbf{Acknowledgments} This work was partially supported by 0311/SBAD/0740. M. Lango was supported by 0311/SBAD/0743.

\bibliographystyle{plain}
\bibliography{jmlr-sample}

\end{document}